\pdfoutput=1

\documentclass[11pt]{article}

\usepackage{EMNLP2023}
\usepackage{times}
\usepackage{latexsym}
\usepackage{graphicx}
\usepackage[utf8]{inputenc} 
\usepackage[T5]{fontenc}    
\usepackage{vntex}
\usepackage{amsmath}

\usepackage[utf8]{inputenc}

\usepackage{microtype}

\usepackage{inconsolata}

\usepackage{graphicx}
\usepackage{tikz}
\usetikzlibrary{positioning}
\usepackage{subcaption}
\usepackage{microtype}
\usepackage[normalem]{ulem}
\usepackage{adjustbox}
\usepackage{multirow}
\usepackage{tabularx}
\usepackage{makecell}
\usepackage{etoc}
\usepackage[accsupp]{axessibility}
\usepackage{xspace}
\usepackage{diagbox}
\usepackage{wrapfig}
\usepackage{bbding}
\usepackage{pifont}
\usepackage{wasysym}
\usepackage{hyperref}

\title{LaVy: Vietnamese Multimodal Large Language Model}
\author{Chi Tran \\  \scalebox{0.9}{Hanoi University of Science and Technology}  \\ \texttt{chi.tb200083@sis.hust.edu.vn} \\\And
Huong Le Thanh \\  \scalebox{0.9}{Hanoi University of Science and Technology} \\ \texttt{huonglt@soict.hust.edu.vn}}

\begin{document}
\renewcommand{\tablename}{Table}
\renewcommand{\figurename}{Image}
\maketitle
\begin{abstract}
Large Language Models (LLMs) and Multimodal Large language models (MLLMs) have taken the world by storm with impressive abilities in complex reasoning and linguistic comprehension. Meanwhile there are plethora of  works related to Vietnamese Large Language Models, the lack of high-quality resources in multimodality limits the progress of Vietnamese MLLMs. In this paper, we pioneer in address this by introducing  LaVy, a state-of-the-art Vietnamese MLLM, and we also introduce LaVy-Bench benchmark designated for evaluating MLLMs's understanding on Vietnamese visual language tasks. All code and model weights are public at https://github.com/baochi0212/LaVy  
\end{abstract} 

\section{Introduction}
In recent years, Large Language Models (LLMs) have demonstrated remarkable capabilities in various natural language processing tasks, showcasing their proficiency in complex reasoning and linguistic comprehension. The success of LLMs has inspired researchers to explore the potential of Multimodal Large Language Models (MLLMs), which incorporate visual information alongside textual data. MLLMs have shown promising results in tasks that require understanding the interplay between language and vision, such as image captioning, visual question answering, and multimodal machine translation.

While there has been significant progress in developing Vietnamese LLMs, the lack of high-quality multimodal resources has hindered the advancement of Vietnamese MLLMs. The availability of diverse and well-annotated datasets is crucial for training and evaluating MLLMs, as they rely on the integration of visual and textual information to perform multimodal tasks effectively.

To address this limitation and foster research in Vietnamese multimodal language understanding, we introduce LaVy, Vietnamese first MLLM and achieve state-of-the-art performance in Vietnamese vision language tasks. LaVy is designed to leverage the rich visual and linguistic information present in Vietnamese data, enabling it to tackle a wide range of multimodal tasks with improved performance. Our model outperforms a multilingual baseline  mBLIP \cite{mblip} on different tasks by a large margin. By developing LaVy, we aim to bridge the gap between Vietnamese LLMs and MLLMs, providing researchers and practitioners with a powerful tool for exploring the intersection of language and vision in the Vietnamese context.

Furthermore, to facilitate the evaluation and comparison of Vietnamese MLLMs, we propose the LaVy-Bench benchmark. This benchmark consists an open VQA task and an in-the-wild test set, specifically designed to assess the visual language understanding and generation capabilities of MLLMs in the Vietnamese and in-the-wild images. By establishing a standardized evaluation framework, we aim to promote the development and benchmarking of Vietnamese MLLMs, driving innovation and collaboration within the research community.

In this paper, we present LaVy and the LaVy-Bench benchmark as significant contributions to the field of Vietnamese multimodal language understanding. We provide a detailed description of LaVy's architecture, data curation and training procedure. Additionally, we introduce the LaVy-Bench benchmark, discussing its design principles, task composition, and evaluation metrics. Through extensive experiments and analysis, we demonstrate the effectiveness of LaVy and the utility of the LaVy-Bench benchmark in advancing Vietnamese MLLM research. 

\section{Related work}
\subsection{Large Language Model}
Recent advancements in Large Language Models (LLMs) have showcased remarkable capabilities across various natural language processing tasks, including dialogue, creative writing, and problem-solving. Models such as  LLaMA \cite{touvron2023llama, touvron2023llama2}, Mistral \cite{mistral}, and Gemma \cite{gemma} have leveraged scalable Transformer-based architectures \cite{transformer} and large-scale  data to become foundation models for general reasoning tasks. These models have demonstrated impressive performance and have set new benchmarks in the field.

Following the trend of LLMs, several Vietnamese language models emerged such as  PhoGPT \cite{phogpt},  Vistral \cite{vistral} perform outstandingly   in Vietnamese LLM benchmarks and NLP tasks. 
\subsection{Multimodal Large Language Model}
Witness the exceptional performance of GPT-4 \cite{gpt4} and Gemini Pro Vision \cite{team2023gemini} in visual language tasks, recent research has focused on developing Multimodal Large Language Models (MLLMs) to achieve unified understanding and reasoning across different modalities, building upon the success of Large Language Models (LLMs). Various methods have been proposed to integrate information from multiple modalities into pre-trained LLM architectures. For instance, Flamingo \cite{alayrac2022flamingo} and BLIP-2 \cite{blip2} introduce different techniques for fusing visual tokens with frozen LLMs using gated attention or Q-former. Inspired by the effectiveness of instruction tuning, LLaVA \cite{liu2023llava} and MiniGPT-4 \cite{zhu2023minigpt} align visual input with LLMs through visual instruction tuning, demonstrating impressive results. Another active line of work is researching efficient MLLMs, resulting in lightweight model families like Bunny \cite{bunny}. Meanwhile, recent works are pioneering development in vision-language tasks for low-resource languages, such as Peacock \cite{peacock}

\section{LaVy}
\subsection{Architecture}
Our model are built with LlaVA architecture \cite{liu2023llava} using three main components:
\begin{itemize}

\item \textbf{Vision Encoder:} The CLIP-Large model \cite{clip}, is used as the vision encoder. 

\item \textbf{MLP Projector:} A two-layer Multi-Layer Perceptron (MLP) projector is employed to align the output representations from the visual and language modalities. This projector ensures that the visual and textual information is transformed into a common space.

\item \textbf{Language Model:} The third component is a Large Language Model, which is a language model responsible for generating textual information, 
take the aligned representations from the MLP 
projector.
\end{itemize}

\subsection{Data Curation}
The barrier for Vietnamese MLLMs' development is resource for training, which is handled by the our novel pipeline of data collection.  
\begin{itemize}
    \item \textbf{Translated then Refined:} 
     We utilize LlaVA training data composing of filtered 558K LAION-CC-SBU captions and 150K GPT-generated multimodal instructions. With the acknowledgement of the insufficient competency of open-source translation projects and translation API like VinAI Translate \cite{vinaitranslate}, Google Translate, ... in convert English data to Vietnamese high-quality data,  we firstly translate a sample with VinAI translation, then we prompt Gemini Pro to rewrite it in more accurate and natural way for Vietnamese language by pairing translated sample and original English sample in Gemini prompt. Because the captions dataset is massive, we only refine a subset of 150K randomly sampled captions, and combine it with 558K non-write captions to finalize a  captioning dataset. Finally, GPT-generated instructions is all refined to make 150K high-quality Vietnamese instructions  
    \item \textbf{Synthetic:} 
    Understanding the difference between Vietnamese images and LlaVA's images, we crawl 8.000  images from web in various topics (for example: Vietnamese event images with keyword \textit{Ảnh sự kiện Việt Nam}) and prompt Gemini Pro Vision to generated concise and detailed description of them to enhance LaVy's performance in Vietnamese images. Totally, we have 16.000 Vietnamese descriptions for crawled images and merge them with rewritten instructions 
\end{itemize}
In eventuality, we curated Vietnamese datasets with 708K image-caption pairs  for pretraining and 166K high-quality instructions for finetuning in training step. Our pipeline is clearly illustrated in Image 1.

\begin{figure}[]
\includegraphics[scale=0.24]{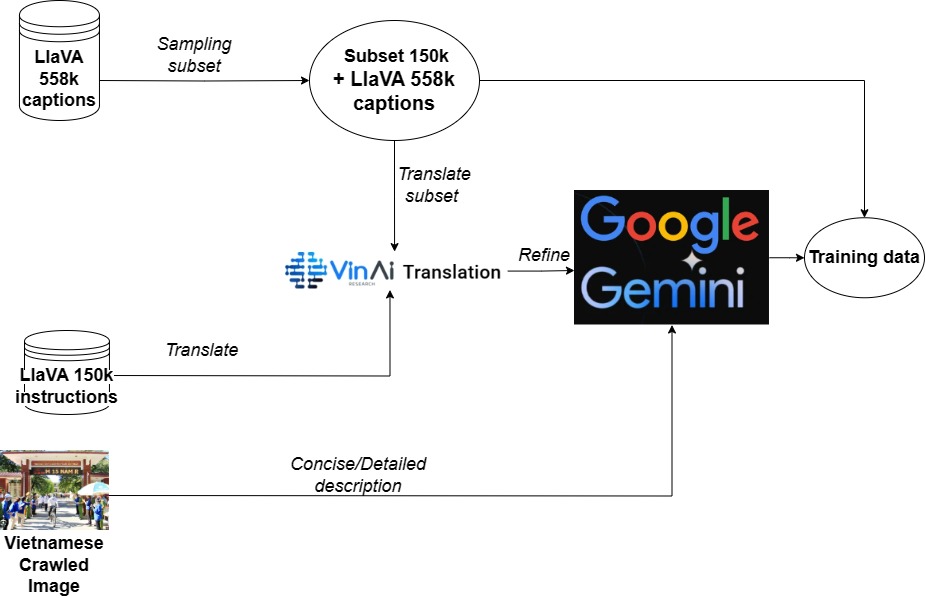}
\caption{Data pipeline}
\end{figure}

\subsection{Training Procedure}
The training procedure is divided into 2 steps:
\begin{itemize}

\item  \textbf{Pretraining:}  Align the vision embeddings from a pre-trained vision encoder with the text embeddings from the LLM by optimizing only the cross-modality projector using a cross-entropy loss for next token prediction. 

\item \textbf{Finetuning:} We apply visual instruction tuning to fully utilize the MLLM's capabilities across different multimodal tasks. We use the same cross-entropy loss as in the pre-training stage, but this time, they employ Low-Rank Adaptation (LoRA) to train both the cross-modality projector and the LLM backbone.

\end{itemize}
\section{Experiment}
\subsection{Implementation details}
We use Vistral 7B as LLM backbone and CLIP large visual encoder.
The training process for the LaVy consists of two stages. In the first stage, the model is pretrained using a dataset of 708k captions for 1 epoch, with a global batch size of 64 and a learning rate of 1e-3. During this stage, all model parameters are frozen, except the MLP layers. Besides, we don't shuffle data but train the model to learn from unrefined data to refined data.  

The second stage involves finetuning the model using an instructions dataset. This stage also spans 1 epoch, with a global batch size of 32 and a learning rate of 2e-5. In this phase, only the newly introduced LoRA (Low-Rank Adaptation) parameters are trainable. \\
Besides, in evaluation, we apply greedy decoding to generate all models' responses 
\subsection{LaVy-Bench}
We construct LaVy-Bench to benchmark models' Vietnamese Visual Language understanding. We use mBLIP models as multilingual baselines. Additionally, we compare performance with close source projects Gemini Pro Vision and SeaLLLM \footnote{\url{https://huggingface.co/spaces/SeaLLMs/SeaLLM-7B}}

\subsubsection{Zero-shot Visual Question Answering (VQA)}
We evaluate the zero-shot Visual Question Answering (VQA) performance of models on the OpenViVQA \cite{openvivqa} dev set, which consists of 3,505 samples. This dataset challenges the models' understanding of the relationships between Vietnamese images and natural language. Furthermore, we propose a new metric for automatic evaluation to replace older metrics, such as BLEU \cite{bleu}, which do not accurately reflect models' competency in the VQA tasks. Our metric is inspired by LLM-as-a-Judge \cite{llm-judge}, which utilizes Gemini Pro to verify the accuracy of generated responses for question-answer pairs.
In Table 1, it's apparent that LaVy's zero-shot VQA performance (\text{45.3\%}) outshadows mBLIP-Bloomz-7B (\text{27.9\%}) and mBLIP-mT0-XL-5B (\text{20.0\%}). However, roughly half of the OpenViVQA dataset is composed of TextQA samples (1,733 samples) that do not appear in our training dataset, making OpenViVQA particularly challenging for our model, not mention to our training instructions just includes descriptions of 8.000 Vietnamese crawled images. Notably, Gemini Pro (Vision) obviously  showcase the best performance, but only correctly answers \text{66.8\%} questions.  
\begin{table}[t]
\centering
\begin{tabular}{|l|c|}
\hline
\textbf{Model} & \textbf{Accuracy} \\
\hline 
\textbf{mBLIP (mT0-XL-5B)} & 20.0 \\
\textbf{mBLIP (BLOOMZ-7B)} &  27.9 \\
\textbf{LaVy} & \underline{45.3} \\
\textbf{Gemini Pro-20240526} & \textbf{66.8} \\
\hline
\end{tabular}
\caption{Zero-shot VQA on OpenViVQA dev set. Models' output accuracy are evaluated by Gemini Pro}
\label{tab:model_accuracy}
\end{table}

\subsubsection{In-the-wild benchmark}
 To further assess the models' comprehension, we follow the evaluation methodology  LLaVA benchmark (in-the-wild) \cite{liu2023llava} and recollect set of 24 diverse images and 60 questions in 3 main types: Complex Reasoning, Detail Description and Conversation. The collected images and manually crafted questions aim to diversify the test set in various aspects: culture, race, image types,... and avoiding opting for images in crawled training images. For each question, we then prompt Gemini Pro Vision to generate detailed description of image and answer the question, and finally use Gemini Pro to rate the models' response on the scale of 1 to 5 while taking Gemini Pro's  response as ground-truth. We opted against using detailed descriptions as references (unlike LlaVA) due to their occasional irrelevance. Instead, we found Gemini Pro Vision's in-the-wild benchmark responses to be closely pertinent. Then we ask model to score outputs in 3 criteria: relevancy, accuracy and naturalness, and sum them all for final score at the end \\
 In comparison with mBLIP baselines in Table 2, LaVy outperforms sharply in all types of questions: Conversation (+\text{42.8\%}), Detail Description (+\text{72.4\%}) and  Complex Reasoning (+\text{50.9\%}). In overall, our model is scored \text{67.2} by Gemini Pro. Further, Our model is only marginalized by SeaLLLM in Detail aspect, and scored higher by Gemini Pro (\text{67.2\%} vs \text{65.9\%}). Some qualitative test cases are depicted in Table 3.

\begin{table}[h]
\begin{tabular}{l|ccc|c}
\hline

 \textbf{Model} & \textbf{Conversation} & \textbf{Detail description} & \textbf{Complex reasoning  }& \textbf{All} \\ 
 \hline
\textbf{mBLIP (mT0-XL-5B)} & 48.5  & 30.8  & 28.9 & 36.1 \\
\textbf{mBLIP (BLOOMZ-7B)}  & 54.2  & 30.8  & 47.1 & 44.0 \\  
\textbf{LaVy} & \textbf{77.4}  & \underline{53.1} & \textbf{71.1}  &  \textbf{67.2} \\
\textbf{SeaLMMM-20240526} & 72.3  &     \textbf{55.3} & 70.1  &  65.9 \\
\hline
\end{tabular}
\caption{Performance on in-the-wild benchmark}
\label{tab:table_label}
\end{table}

\section{Limitations}
Our model still have several limitations: 
\begin{itemize}
    \item Although LaVy exhibits deep understanding in Vietnamese visual language tasks but still faces many challenges, for example: TextQA, due to lack of high-quality annotated data for these tasks.
    \item Moreover, like other MLLMs, our model still suffers from hallucination where it generates irrelevant information, redundant details or misinformation.

\end{itemize}
\section{Conclusion}
In this paper, we have introduced LaVy, a pioneering state-of-the-art Vietnamese Multimodal Large Language Model (MLLM) that aims to address the lack of high-quality resources in multimodality for the Vietnamese language. LaVy represents a significant step forward in the development of Vietnamese MLLMs, enabling complex reasoning and linguistic comprehension in tasks that involve both visual and textual information.

Furthermore, we have presented LaVy-Bench, a comprehensive benchmark designed specifically for evaluating the performance of MLLMs on Vietnamese visual language tasks. This benchmark provides a standardized platform for assessing the capabilities of Vietnamese MLLMs, facilitating the comparison and advancement of these models. Our model also have proved SOTA perfomance in comparison with mBLIP baselines in test sets of benchmark.

As future work, we plan to expand the capabilities of LaVy by incorporate diverse instructions to entirely handle challenging tasks like Vietnamese OCR, Object Counting. We hope our work will contribute to the advancement of Vietnamese MLLMs' development.
\newcommand\HumanIcon{\raisebox{0.3ex}{\tikz\fill[scale=0.4] (0,0) circle (1ex);}}
\newcommand\BunnyIcon{\tikz\draw[scale=0.4,circle,fill=gray] (0,0) circle (1ex);}
\begin{table*}[h]
 \small
  \centering
  \begin{tabular}{  m{3cm} | m{12cm}  }
    \hline
    \begin{minipage}[t]{\linewidth}
		\centering
		\vspace{-10.3ex} 
		\includegraphics[width=0.9\linewidth]{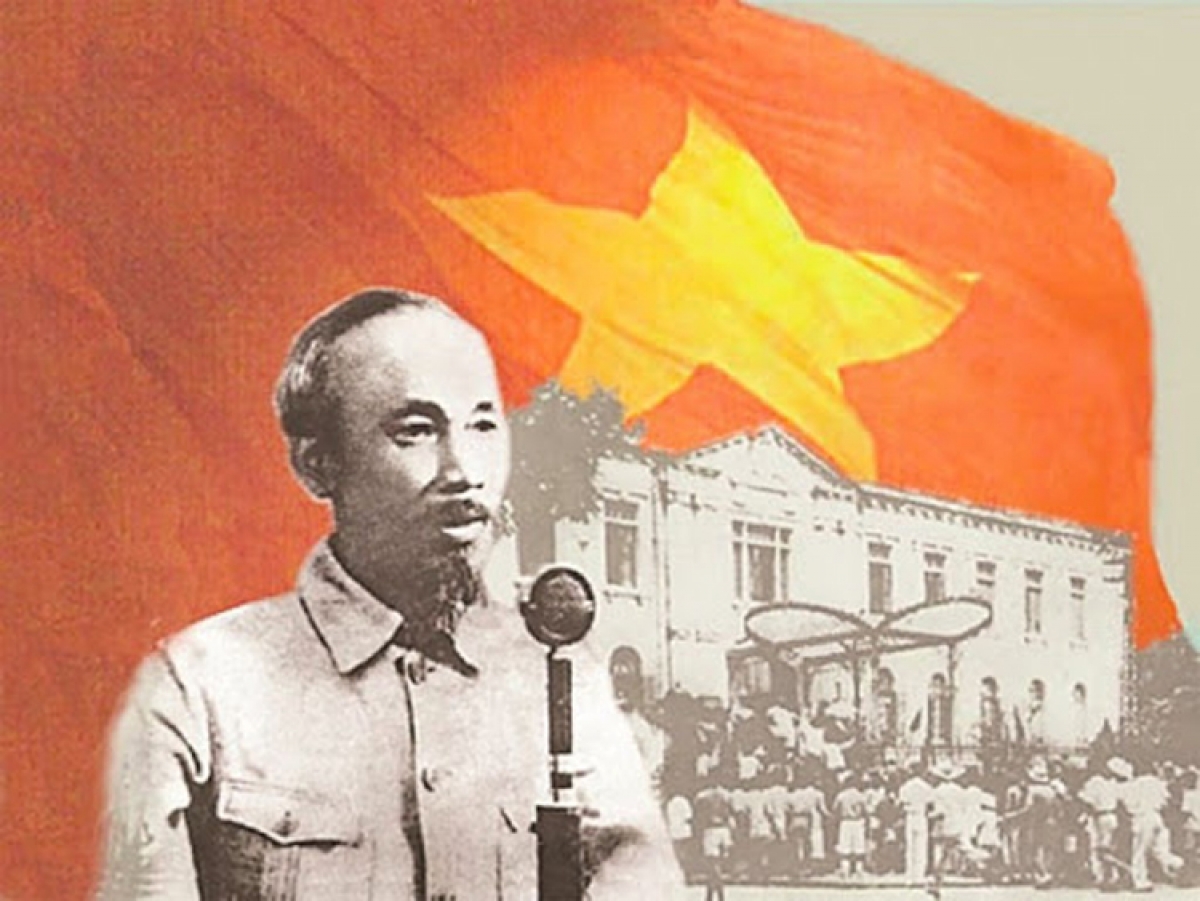}
		\vspace{0.7ex}
	\end{minipage}
    & \vspace{1ex} {\HumanIcon} \textbf{Đây là nhân vật lịch sử Việt Nam nổi tiếng nào?}\newline Hình ảnh cho thấy Hồ Chí Minh, một nhà lãnh đạo cách mạng và chính trị gia Việt Nam. Ông là người sáng lập Đảng Cộng sản Việt Nam và là Chủ tịch nước Việt Nam Dân chủ Cộng hòa từ năm 1945 đến khi qua đời vào năm 1969. Hồ Chí Minh là một nhân vật quan trọng trong lịch sử Việt Nam, đóng vai trò then chốt trong cuộc đấu tranh giành độc lập và thống nhất đất nước.\newline
    {\HumanIcon} \textbf{Bức ảnh có ý nghĩa như thế nào với lịch sử Việt Nam?}\newline Bức ảnh đen trắng này có ý nghĩa quan trọng trong lịch sử Việt Nam vì nó ghi lại khoảnh khắc lịch sử khi Hồ Chí Minh, lãnh tụ của Việt Nam, đang phát biểu trước đám đông. Bức ảnh cho thấy Hồ Chí Minh đang đứng trên bục phát biểu, cầm micro và nói chuyện với mọi người. Sự kiện này có thể là một cuộc mít tinh hoặc một bài phát biểu quan trọng, thể hiện tầm ảnh hưởng và vai trò lãnh đạo của Hồ Chí Minh trong phong trào độc lập và tự do của Việt Nam. Bức ảnh này là một minh chứng trực quan cho những đóng góp của Hồ Chí Minh đối với lịch sử Việt Nam và di sản của ông vẫn tiếp tục được tôn vinh và ngưỡng mộ cho đến ngày nay.\newline \\
    \hline
        \begin{minipage}[t]{\linewidth}
        \centering
        \includegraphics[width=0.9\linewidth]{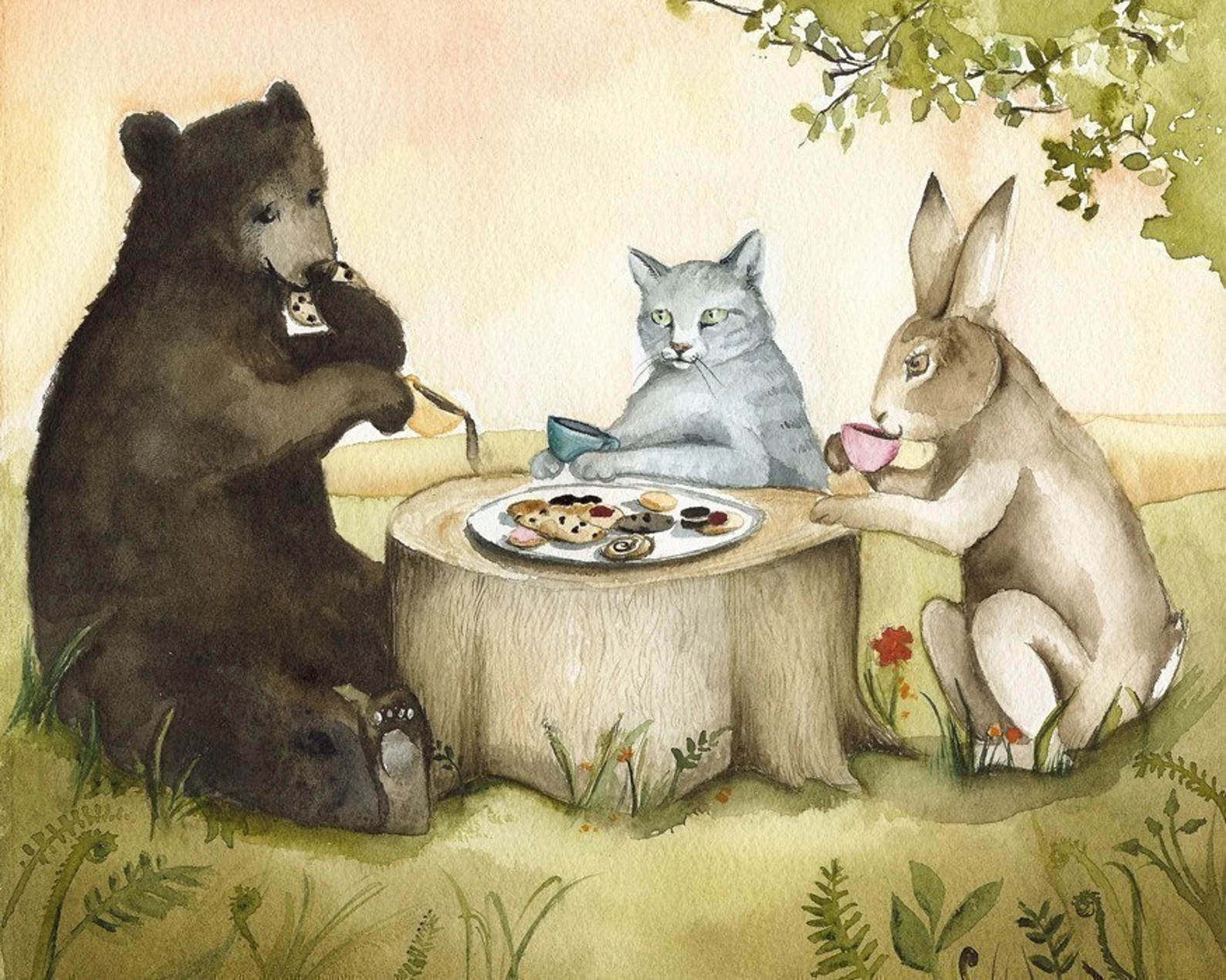}
    \end{minipage}
    & {\HumanIcon} \textbf{Các con vật nào xuất hiện trong ảnh, và chúng đang làm gì}\newline Trong ảnh có một con gấu, một con mèo và một con thỏ. Chúng đang ngồi quanh bàn ăn, cùng nhau thưởng thức bữa ăn.\newline
    {\HumanIcon} \textbf{Viết một câu chuyện cổ tích cho bức ảnh trên.}\newline Ngày xửa ngày xưa, trong một khu rừng xanh tươi, có một chú gấu nâu, một chú mèo xám và một chú thỏ trắng cùng nhau dùng bữa. Họ ngồi quanh một chiếc bàn gỗ nhỏ, trên bàn bày nhiều món ăn ngon. Chú gấu và chú mèo ngồi đối diện nhau, còn chú thỏ ngồi giữa họ. Khi họ thưởng thức bữa ăn, chú gấu và chú mèo trò chuyện vui vẻ, còn chú thỏ chăm chú lắng nghe. Họ nói về những cuộc phiêu lưu trong rừng, những câu chuyện về gia đình và những điều kỳ diệu của thiên nhiên. Khi bữa ăn kết thúc, họ cảm thấy gắn bó và thân thiết hơn. Họ hứa sẽ luôn giúp đỡ nhau và cùng nhau khám phá khu rừng. Từ ngày đó, họ trở thành những người bạn tốt, cùng nhau chia sẻ những câu chuyện và cuộc phiêu lưu trong thế giới tự nhiên. \\ \hline
    
    \begin{minipage}[t]{\linewidth}
        \centering
        \vspace{-10ex} 
        \includegraphics[width=0.9\linewidth]{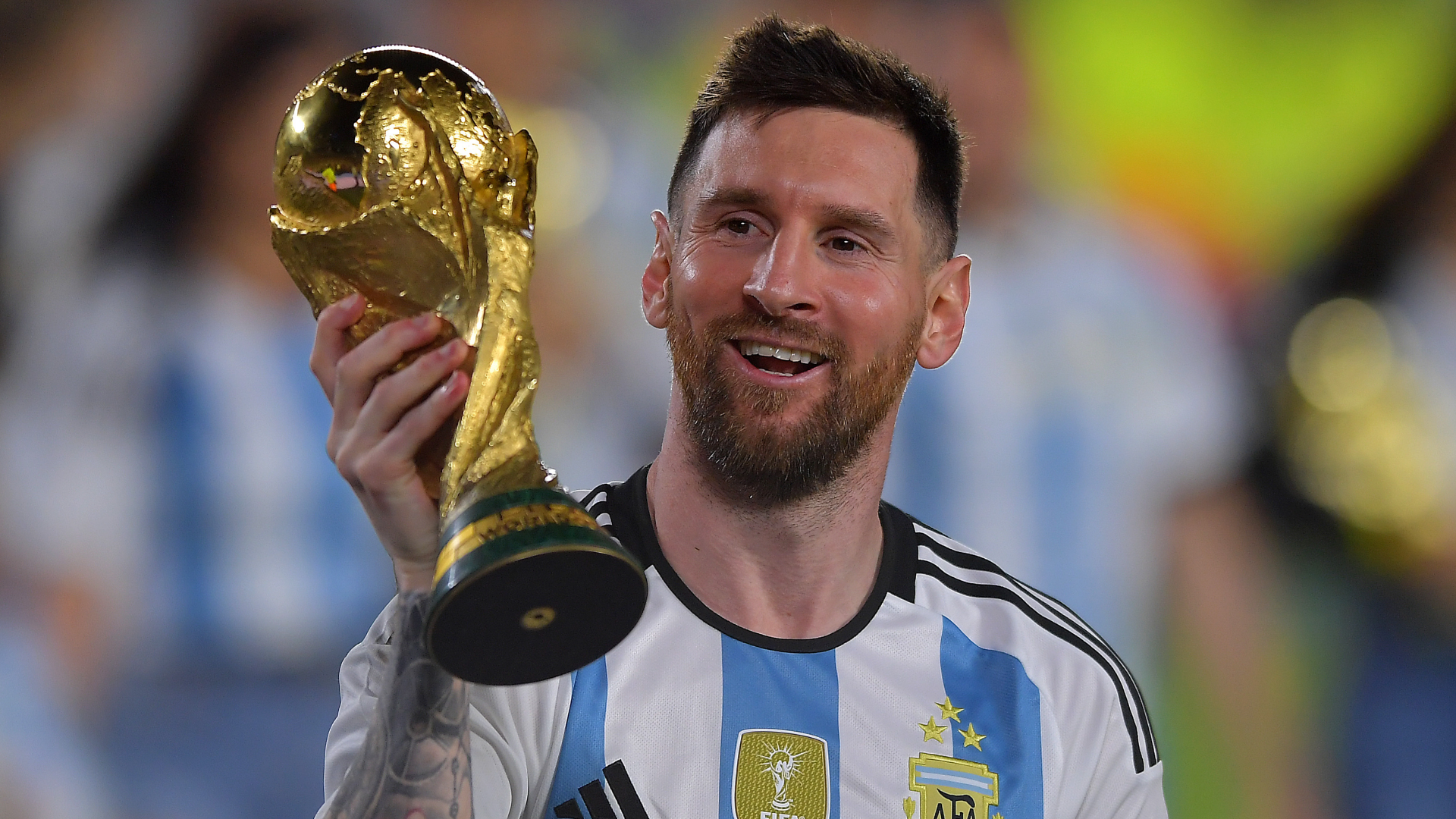}
        \vspace{0.7ex} 
    \end{minipage}
    & \vspace{3ex} 
    {\HumanIcon} \textbf{Bức ảnh được chụp ở sự kiện nào?}\newline Bức ảnh được chụp tại lễ trao giải FIFA World Cup, nơi Lionel Messi đang cầm chiếc cúp vàng.\newline
    {\HumanIcon} \textbf{Bạn có biết tên của người đàn ông trong bức ảnh là gì?}\newline Người đàn ông trong bức ảnh là Lionel Messi.\newline
    {\HumanIcon} \textbf{Miêu tả cảm xúc của người đàn ông trong bức hình}\newline Người đàn ông trong bức hình đang rất vui mừng và hạnh phúc. Anh ta đang cầm trên tay chiếc cúp vàng World Cup và nở một nụ cười rạng rỡ. Có thể thấy rằng anh ta đang rất tự hào về thành tích của mình và cảm thấy vô cùng phấn khích.\\ \hline
    
    \begin{minipage}[t]{\linewidth}
        \centering
        \includegraphics[width=0.9\linewidth]{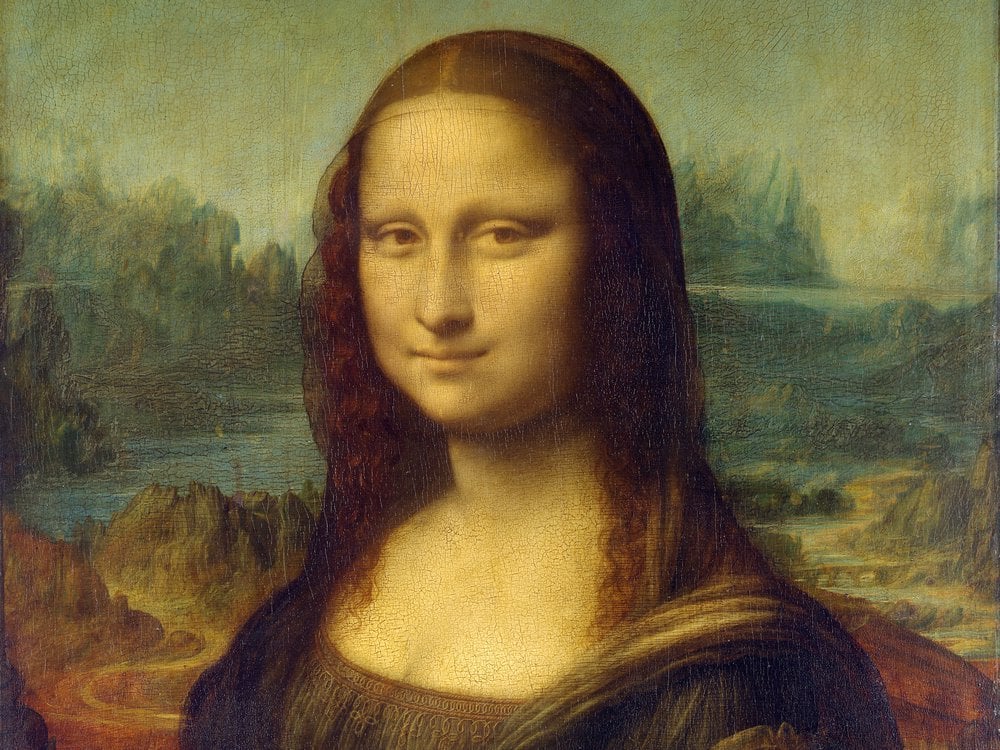}
    \end{minipage}
    & {\HumanIcon} \textbf{Bạn có biết ai vẽ bức tranh này không}\newline Đây là bức tranh Mona Lisa của Leonardo da Vinci.\newline
    {\HumanIcon} \textbf{Thảo luận về tác động lịch sử và ý nghĩa của bức tranh này trong thế giới nghệ thuật.}\newline Bức tranh Mona Lisa của Leonardo da Vinci, với nụ cười bí ẩn và đôi mắt hướng về phía người xem, là một tác phẩm nghệ thuật mang tính biểu tượng và có ảnh hưởng lớn trong thế giới nghệ thuật. Được vẽ vào đầu thế kỷ 16, bức tranh này thể hiện kỹ năng hội họa bậc thầy của da Vinci, với sự chú ý đến từng chi tiết và cách sử dụng ánh sáng và bóng tối tinh tế. Bức tranh đã trở thành một trong những tác phẩm nghệ thuật nổi tiếng và được yêu thích nhất trên thế giới, thu hút hàng triệu du khách đến Bảo tàng Louvre ở Paris mỗi năm. Tác động lịch sử của bức tranh này nằm ở sự đổi mới nghệ thuật và ảnh hưởng của nó đối với các nghệ sĩ trong nhiều thế kỷ. Bức tranh đã truyền cảm hứng cho vô số nghệ sĩ và nhà sử học nghệ thuật, những người đã nghiên cứu và phân tích các kỹ thuật, bố cục và biểu đạt cảm xúc của nó. Bức tranh cũng đã được sao chép, tái tạo và diễn giải vô số lần, cho thấy sức hấp dẫn lâu dài và tác động của nó đối với thế giới nghệ thuật. Ngoài ra, bức tranh còn có ý nghĩa lịch sử vì nó là một trong những tác phẩm nghệ thuật đầu tiên được bảo vệ nghiêm ngặt, với các biện pháp an ninh được tăng cường để bảo vệ bức tranh khỏi bị hư hại hoặc trộm cắp. Sự nổi tiếng và tầm quan trọng của bức tranh Mona Lisa đã khiến nó trở thành một biểu tượng văn hóa và là một phần không thể thiếu trong di sản nghệ thuật của nhân loại.\newline\\ \hline
    
    \begin{minipage}[t]{\linewidth}
        \centering
        \includegraphics[width=0.9\linewidth]{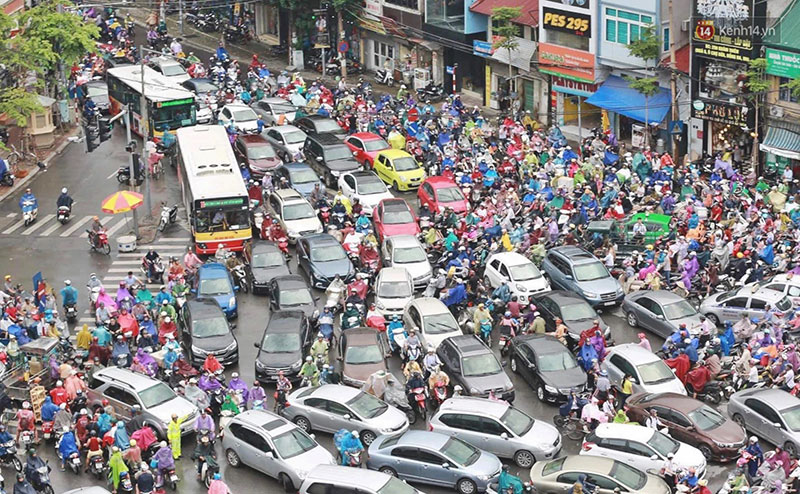}
    \end{minipage}
    & {\HumanIcon} \textbf{Bức ảnh phản ánh vấn đề xã hội gì? Có những giải pháp nào để giải quyết vấn đề này?}\newline Bức ảnh cho thấy một con phố đông đúc với nhiều người đi bộ, xe máy và ô tô. Tình trạng tắc nghẽn giao thông và đông đúc này có thể gây ra nhiều vấn đề xã hội như ô nhiễm không khí, tiếng ồn, tai nạn và căng thẳng cho người tham gia giao thông. Để giải quyết những vấn đề này, có thể áp dụng nhiều giải pháp như cải thiện cơ sở hạ tầng giao thông công cộng, khuyến khích đi bộ và đi xe đạp, thực hiện các biện pháp quản lý giao thông hiệu quả và thúc đẩy các phương thức di chuyển thay thế như đi chung xe hoặc đi chung xe. Ngoài ra, việc nâng cao nhận thức của người dân về tác động môi trường của giao thông và khuyến khích các hành vi thân thiện với môi trường cũng có thể góp phần giảm ùn tắc giao thông và cải thiện chất lượng cuộc sống chung trong thành phố.\\ \hline
    
  \end{tabular}
  \caption{ Qualitative test cases of LaVy. In-the-wild images and questions are varied in image type, topic, culture...}
\label{fig:demo}
\end{table*}
\clearpage
\bibliographystyle{acl_natbib}
\bibliography{main}
\clearpage

\end{document}